\documentclass[conference]{IEEEtran}
\IEEEoverridecommandlockouts
\renewcommand{\thesection}{\Roman{section}}

\usepackage{booktabs}
\usepackage{fancyhdr}
\usepackage{algorithm}
\usepackage[usenames, dvipsnames]{xcolor}
\usepackage{algorithmic}
\usepackage{colortbl}
\usepackage{caption}
\usepackage{array}
\usepackage{adjustbox}
\usepackage{cite}
\usepackage{hyperref,graphicx,color,float}
\usepackage{multirow}
\usepackage{amsmath,amssymb,amsfonts}
\usepackage{textcomp}
\usepackage{comment}

\definecolor{myblue}{RGB}{10, 150, 200}
\definecolor{highlightColor}{HTML}{E6FFE6}

\def\BibTeX{{\rm B\kern-.05em{\sc i\kern-.025em b}\kern-.08em
    T\kern-.1667em\lower.7ex\hbox{E}\kern-.125emX}}

\fancypagestyle{firstpage}{
  \fancyhf{}
  \fancyhead[L]{\small 2026 IEEE International Conference on Optics, Machine Learning
    and Emerging Technology (OMLET) \\ 29 -- 31 October 2026, Nairobi, Kenya}
  \fancyfoot[L]{\small 979-8-3195-1287-1/26/\$31.00 \copyright2026 IEEE}

}
\title{Frequency-Domain AI-Generated Image Detection: Exploring Decoder and Channel Attention for Feature Refinement}

\author{
  \IEEEauthorblockN{%
    Uday Shankar Roy\textsuperscript{1},
    Mahbuba Jahan Minu\textsuperscript{2}}
  \IEEEauthorblockA{%
    \textsuperscript{1,2}Computer Science and Engineering, Khulna University of Engineering \& Technology\\
    Khulna, Bangladesh\\
    \textsuperscript{1}udayroycse@gmail.com,
    \textsuperscript{2}mjm1807120@gmail.com}
}

\begin{document}
\maketitle
\thispagestyle{firstpage}

\begin{abstract}
With the rapid progress of AI, the number of AI-generated images has increased
significantly in recent years. However, the increasing variety of image generation models makes
detection more difficult. In this work, we use Fast Fourier Transform (FFT) representation with
EfficientNet-B0 for AI-generated image detection. EfficientNet-B0 provides a lightweight
architecture that can be useful for resource-limited applications. Most frequency-domain
detectors use a standard encoder to extract features from the FFT spectrum and directly pass
them to a classifier. We explored a different approach by investigating ECA, U-Net, and
Attention U-Net as alternatives to this direct encoder-to-classifier approach. ECA applies
channel attention, while U-Net and Attention U-Net use decoder-based architectures to recover
and refine spatial information in the extracted frequency features. We used a balanced subset of
the MS COCOAI dataset that includes AI-generated images from five different models. Three runs were carried out for each experiment, and the average values were recorded. Experimental
results indicate that EfficientNet-B0 obtained an accuracy of 84.64\%, which is 4.50 percentage
points higher than the ResNet-50 baseline reported in the dataset paper. EfficientNet-B0 with \mbox{U-Net} provided a small improvement, achieving an accuracy of 84.85\%, while ECA did not
increase the overall performance. EfficientNet-B0 with Attention U-Net achieved the best overall
performance, with an accuracy of 85.51\% and an ROC-AUC of 92.99\%. This represents an
improvement of 0.87 percentage points in accuracy compared to the EfficientNet-B0 baseline and 5.37
percentage points over the ResNet-50 baseline reported in the dataset paper.
\end{abstract}


\section{Introduction}

AI-generated images have become a real challenge for online platforms, news organizations, and even for ordinary people trying to know whether a photo is genuine or not. Recent advances in image generation have made it possible to produce highly realistic images, creating new challenges for reliable AI-generated image detection. As these images become more realistic, older detection methods are often no longer reliable enough. This has led researchers toward developing automated image detectors that can identify potentially AI-generated content and help prevent misleading or manipulated images from spreading online.

One promising approach is to analyze images in the frequency domain instead of looking only at their pixels. Many generative models rely on upsampling layers to increase image resolution, and these operations can leave subtle patterns in the Fourier spectrum that are not typically found in real photographs. Several groups have shown this independently \cite{frank2020, dzanic2019, durall2020}, and the core finding is the same. The main idea is that the frequency spectrum of a real photograph differs from that of an AI-generated image, and this difference can be used to train a classifier. Frequency-based approaches can be useful, compared with the methods that work directly on pixels, because these spectral patterns may remain detectable even when the generated image looks realistic to people.

To extract useful features from the frequency representation, convolutional neural networks can be used as encoder backbones. EfficientNet-B0 was chosen for its balance of efficiency and accuracy. Its lightweight design makes it suitable for resource-limited applications while still providing sufficient capacity for image classification.

The standard setup for a frequency-domain detector is simple. The image is first converted into its FFT magnitude spectrum, which is then passed through an encoder such as EfficientNet or ResNet. Global average pooling is used to turn the extracted features into a fixed-size vector, followed by a classification head. Though this approach works reasonably well, it can sometimes overlook an important part of the frequency information: its spatial structure across different scales. The earlier layers of the encoder tend to preserve more local and spatial details, whereas deeper layers learn more complex and meaningful features. When all of these features are reduced to a single average at the end, much of this multi-scale information is lost, although it may help differentiate between real and AI-generated images.

This idea led us to explore whether a U-Net-style architecture could also be useful for frequency-domain image classification. U-Net \cite{ronneberger2015unet} was originally developed for medical image segmentation, where preserving spatial information at different scales is important. Its skip connections allow the decoder to recover features from earlier encoder layers while combining them with deeper, more abstract representations. We thought a similar idea could be useful for frequency-based classification, where information from different levels of the feature maps may help distinguish real images from generated ones. Instead of producing a segmentation mask, the decoder can be used to refine and combine the encoder’s spectral features at multiple levels before they are pooled and passed to the classifier. We further extended the U-Net architecture with attention to investigate whether the decoder features could be further improved. In the decoder, Attention U-Net uses spatial and channel attention to focus on important features while reducing less useful information. In addition to the decoder-based approaches, we also investigated Efficient Channel Attention (ECA) to examine whether channel-level attention alone could improve the features extracted by EfficientNet-B0.

To evaluate these approaches, we conducted experiments on a balanced subset of the MS COCOAI dataset containing 16K real and 16K AI-generated images. Each architecture was evaluated using three random seeds, and the average results were reported using Accuracy, Precision, Recall, F1-score, and ROC-AUC.

The key aspects of this work are:
\begin{itemize}
  \item We use EfficientNet-B0 as the baseline encoder for frequency-domain AI-generated image detection. It achieved an accuracy of 84.64\%, which is 4.50 percentage points higher than the ResNet-50 baseline reported in the MS COCOAI dataset paper. Its lightweight architecture also makes it suitable for resource-limited applications.
  \item We explore the use of Efficient Channel Attention (ECA), U-Net, and Attention \mbox{U-Net} for AI-generated image detection in the frequency domain.
  \item We investigate the effect of channel attention and decoder-based feature refinement. The results show that ECA does not improve the performance of the EfficientNet-B0 baseline, while U-Net provides a small improvement.
  \item Attention U-Net obtained the best overall performance among the tested architectures, with an accuracy of 85.51\%, improving the accuracy by 0.87 percentage points over the EfficientNet-B0 baseline.
\end{itemize}

\section{Related Work}

The idea that AI-generated images leave behind spectral fingerprints goes back to a few
key papers published around 2020. Dzanic et al. \cite{dzanic2019} and Frank et al.
\cite{frank2020} both showed, working independently, that the Fourier spectrum of a
generated image looks noticeably different from that of a real photograph, different
enough to build a classifier on. Durall et al. \cite{durall2020} traced this back to how
most generators work: transposed convolution layers produce patterns in the frequency
domain that break the natural 1/f spectral falloff of real images.

Things got more complicated when researchers showed that these artifacts are not always present. Chandrasegaran et al. \cite{chandrasegaran2021} found that a small tweak to
a GAN's upsampling layer is enough to make the spectral artifacts disappear, effectively
breaking frequency-based detectors. Dong et al. \cite{dong2022} repeated the test across
eight architectures and found the same thing. This pushed researchers toward combining spatial and frequency information rather than relying on one alone. Qian et al.
\cite{qian2020} showed that a two-stream model processing both RGB and frequency features
consistently beats either stream on its own, and similar gains have been reported with
wavelet-RGB fusion \cite{wang2023wavelet} and dual-branch transformers
\cite{rozhbayani2025}.

More recently, a group of methods has moved away from memorising what specific generators
look like and toward modelling what real images look like in the frequency domain.
SPAI \cite{karageorgiou2024spai} used a masked spectral pretext task to train a detector
that generalises across 13 generative models, gaining 5.5\% in AUC over the prior state
of the art. FIRE \cite{chu2024fire} reported that diffusion models often fail to capture the mid-frequency patterns of real images during reconstruction and used that gap
as a detection signal. FRADet \cite{li2025fradet} added a residual frequency refinement
step on top of a reconstruction-based detector, reporting a 3.98\% accuracy improvement on
the GenImage benchmark.

On the backbone side, EfficientNet \cite{tan2019efficientnet} has also emerged as a widely used lightweight encoder for frequency-domain pipelines, as it achieves higher accuracy for 
fewer parameters \cite{rathore2025}. Channel attention modules like
ECA \cite{wang2020ecanet} have been added to some pipelines to let the network
emphasise the most informative frequency channels. U-Net \cite{ronneberger2015unet}
introduced skip connections as a way to combine features from multiple encoder depths, a mechanism that has been widely used in segmentation for a decade. To the best of our knowledge, however, using a full U-Net decoder purely as a
multi-scale frequency feature refiner within a classification pipeline has not been
widely explored. Most existing methods pass encoder output directly to a pooling layer,
without using decoder-based feature recovery before classification. This work
investigates whether such a design can improve detection performance.

\section{Methodology}

\begin{figure}[htbp]
  \centering
  \includegraphics[width=\columnwidth]{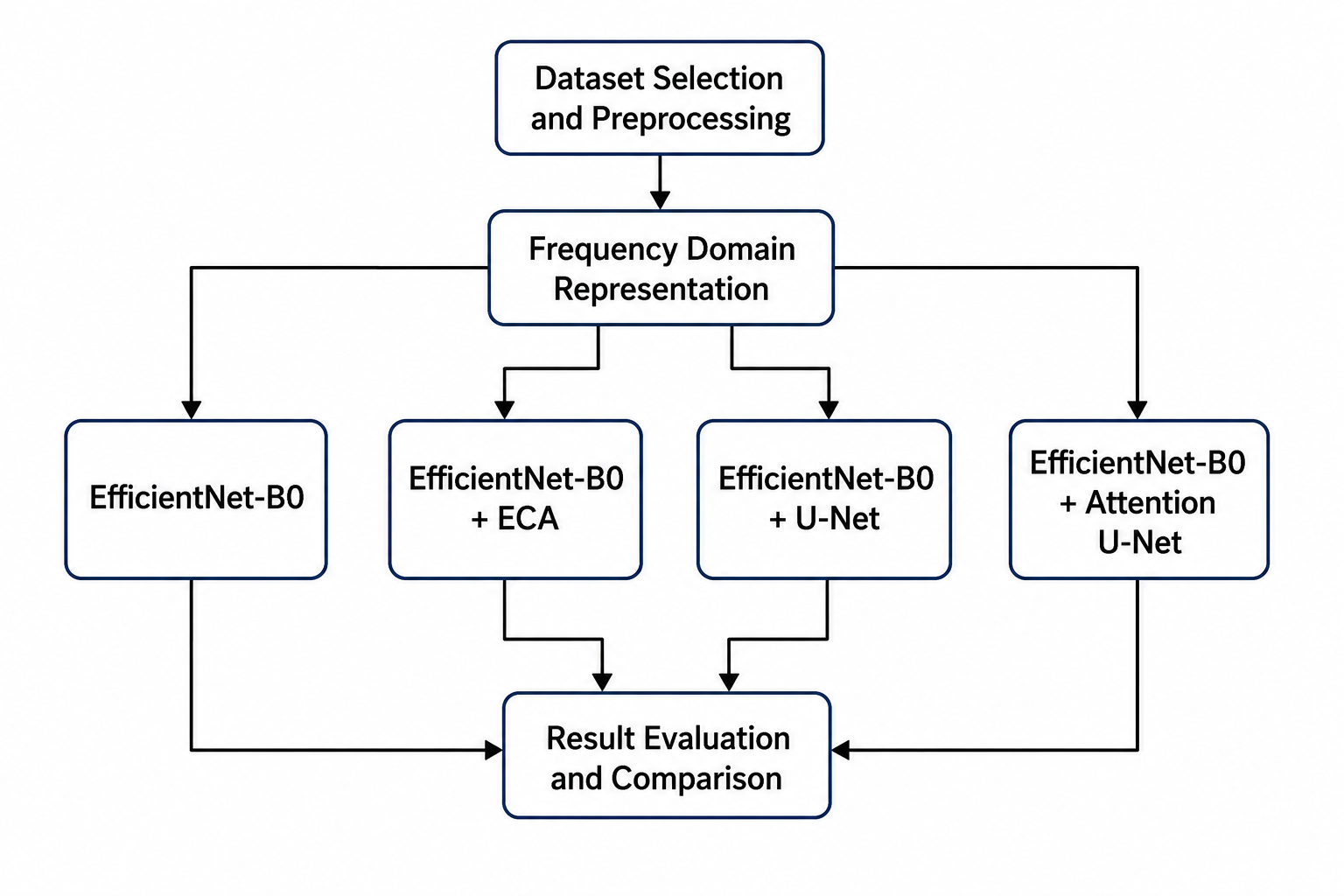}
  \caption{Experimental pipeline}
  \label{fig:pipeline}
\end{figure}

An overview of our proposed work is shown in Fig.~\ref{fig:pipeline}. We used a balanced
subset of the MS COCOAI dataset~\cite{roy2026defactify}. After preprocessing, the images
were converted into frequency-domain representations. 
We first trained the EfficientNet-B0
model as the baseline. We further explored the effectiveness of channel attention (ECA),
U-Net, and Attention U-Net for detecting AI-generated images.

\subsection{Dataset}

The experiments were conducted using the MS COCOAI dataset~\cite{roy2026defactify}. It contains
16K real images and 80K AI-generated images. The 16K real images were collected from the
MS COCO dataset~\cite{lin2014coco}. The human-written caption of each real image was
used to generate five AI-generated images using five state-of-the-art image generation
models: MidJourney v6, Stable Diffusion 2.1, Stable Diffusion 3, DALL-E 3, and SDXL.

\begin{figure}[htbp]
  \centering
  
  \includegraphics[width=0.48\columnwidth]{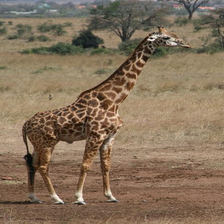}
  \hfill
  \includegraphics[width=0.48\columnwidth]{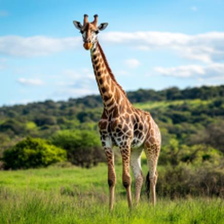}
  \caption{Real (left) and AI-generated (right) sample images from the MS COCOAI dataset~\cite{roy2026defactify}}
  \label{fig:samples}
\end{figure}

Some examples from the dataset are presented in Fig.~\ref{fig:samples}. For our experiments, 16K real images and 16K AI-generated images were selected to create
a balanced dataset of 32K images in total. All images were resized to $224 \times 224$
pixels before being used in the experiments.

\subsection{Frequency-Domain Representation}

The frequency-domain representation of the images was used instead of the original RGB
image. The input RGB image is first converted to grayscale to simplify the frequency
analysis:
\begin{equation}
  I_{\text{gray}} = 0.299\,R + 0.587\,G + 0.114\,B
  \label{eq:gray}
\end{equation}

A two-dimensional Fast Fourier Transform (2D-FFT) is then applied to obtain the
frequency-domain representation:
\begin{equation}
  \mathcal{F}(u,v) = \sum_{x=0}^{M-1}\sum_{y=0}^{N-1}
    I_{\text{gray}}(x,y)\,e^{-j2\pi\!\left(\frac{ux}{M}+\frac{vy}{N}\right)}
  \label{eq:fft}
\end{equation}

The frequency spectrum is shifted to center the zero-frequency component using an FFT
shift operation. The magnitude spectrum is then converted to a logarithmic scale to
compress the large dynamic range of the frequency values:
\begin{equation}
  F(u,v) = \log\!\left(1 + \bigl|\mathcal{F}_{\text{shift}}(u,v)\bigr|\right)
  \label{eq:logmag}
\end{equation}

Finally, min-max normalisation scales the spectrum to $[0,\,1]$:
\begin{equation}
  \hat{F}(u,v) = \frac{F(u,v) - F_{\min}}{F_{\max} - F_{\min}}
  \label{eq:norm}
\end{equation}

As EfficientNet-B0 requires three-channel input, the normalised spectrum $\hat{F}$ is
replicated across three channels before being passed to the network.

\subsection{Baseline Architecture (EfficientNet-B0)}

EfficientNet-B0 \cite{tan2019efficientnet} is used as the baseline architecture for the
frequency-domain image detection task. It is a lightweight model that offers good performance with low computational cost, making it suitable for resource-limited devices such as edge devices. The model uses
pre-trained ImageNet weights. The EfficientNet-B0 feature extractor generates the final feature maps, which are then processed using adaptive average pooling. These features are flattened and passed through a fully connected layer with 256 units. A SiLU activation and 0.5 dropout are applied before the final layer, which produces one output for binary classification.

\subsection{Architectural Variants (ECA, U-Net, Attention U-Net)}

The frequency-domain representation may contain useful information at different levels of
the extracted feature maps. However, the baseline EfficientNet-B0 directly converts the
extracted features into a compact representation for classification. We therefore
investigated different architectural approaches to examine whether the extracted frequency
features could be further improved before classification. ECA, U-Net, and Attention U-Net
were selected for this investigation. 
\subsubsection*{ECA}
Efficient Channel Attention (ECA) \cite{wang2020ecanet} module was tested to investigate
whether channel attention could improve the features extracted by EfficientNet-B0. The
ECA module is applied after the EfficientNet-B0 feature extractor and before the pooling
layer. Global average pooling is first used to summarize each channel. A 1D convolution with a kernel size of 3 then calculates the channel weights, which are passed through a sigmoid function. These weights are applied to the feature maps before they are sent to the classification head.

\subsubsection*{U-Net}
The frequency features extracted by EfficientNet-B0 may lose some spatial information as
the feature maps are progressively downsampled. U-Net \cite{ronneberger2015unet} was used
to investigate whether a decoder could recover and refine this information before
classification. The decoder gradually upsamples the feature maps and combines them with features from the matching encoder stages through skip connections. The decoder produces a 128-channel feature map, which is then processed by adaptive average pooling before going to the classification head.

\subsubsection*{Attention U-Net}
To further investigate whether attention mechanisms could improve the U-Net decoder, an
Attention U-Net variant was tested. Spatial and Channel Squeeze \& Excitation (SCSE)
attention \cite{roy2018scse} was integrated into the decoder stages. These SCSE blocks
apply attention to the decoder feature maps across both spatial and channel dimensions,
allowing the network to selectively emphasise informative frequency features before
classification. 
\begin{figure}[htbp]
  \centering
  \includegraphics[width=\columnwidth]{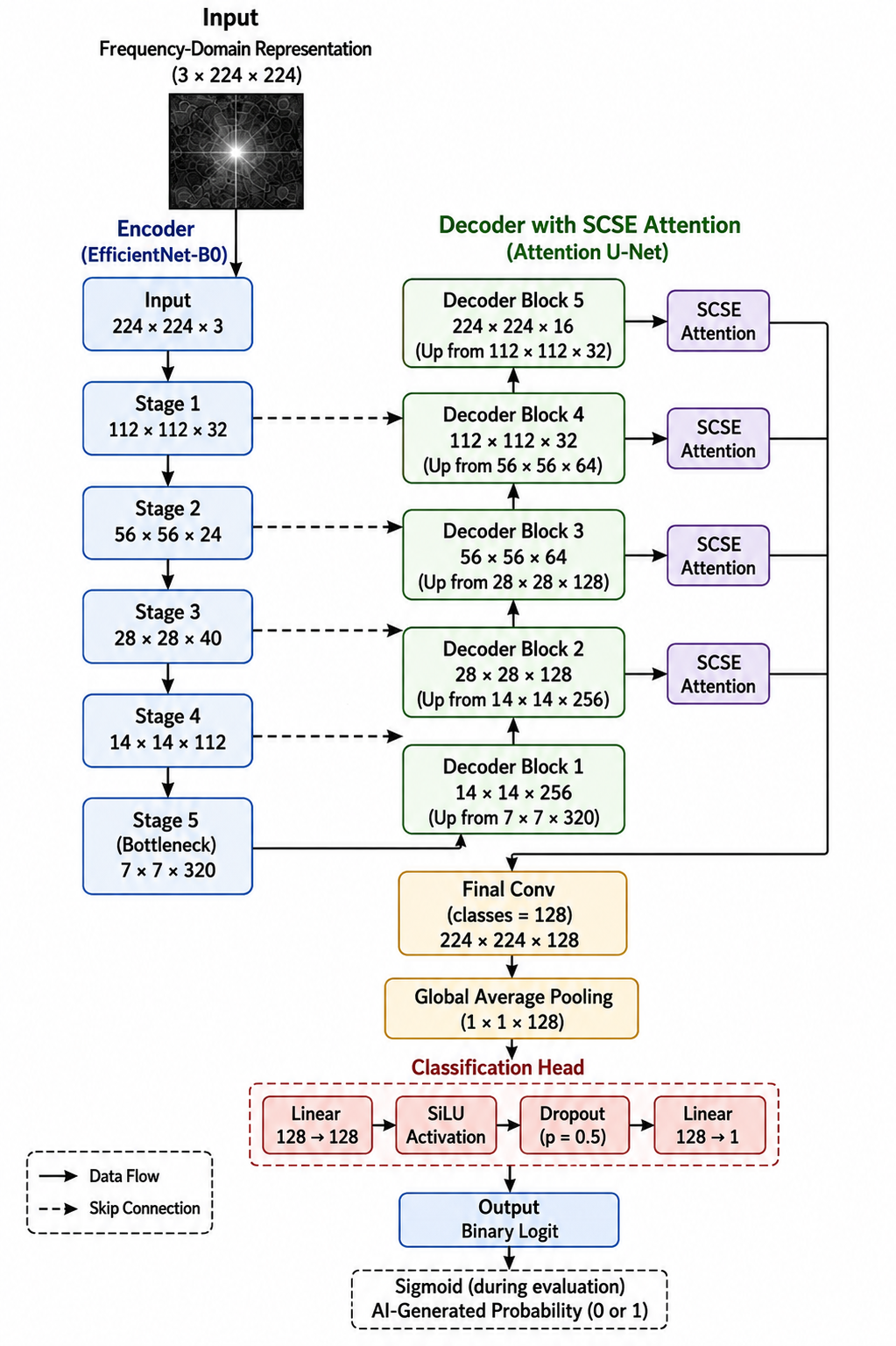}
  \caption{Architecture of the proposed Attention U-Net classifier}
  \label{fig:attn_unet}
\end{figure}
The Attention U-Net decoder outputs a 128-channel feature map, which is reduced through adaptive average pooling before entering a classification head with a 128-unit fully connected layer, SiLU activation, 0.5 dropout, and a binary output layer. The full architecture of the Attention U-Net classifier is illustrated
in Fig.~\ref{fig:attn_unet}.

\subsection{Evaluation Metrics}

Five metrics were considered to assess the classification performance of the models. The output score from the classification head was converted into a probability using the sigmoid function. A threshold of 0.5 was then used to assign each test image to either the real or AI-generated class.

\begin{itemize}
  \item \textit{Accuracy}: represents the proportion of test images that received the correct class label.
  \item \textit{Precision}: quantifies how many of the images predicted as AI-generated
    are actually AI-generated.
  \item \textit{Recall}: indicates the proportion of AI-generated images that were successfully detected by the model.
  \item \textit{F1-Score}: combines Precision and Recall into a single metric using their harmonic mean.
  \item \textit{ROC-AUC}: measures how well the model separates real and AI-generated images across different classification thresholds.
\end{itemize}

All results are averaged across three random seeds (9, 99, 999).

\subsection{Implementation}

The experiments were carried out in Python with PyTorch on Google Colab using a T4 GPU. The models were trained end-to-end with all layers unfrozen. The segmentation\_models\_pytorch (smp) library was used for the U-Net and Attention U-Net variants. The data were split into 80\% training, 10\% validation, and 10\% test sets through stratified sampling. EfficientNet-B0 and ECA were trained with a batch size of 32, whereas a batch size of 16 was used for the U-Net-based models due to their higher memory requirements.

Each experiment was conducted three times with random seeds 9, 99, and 999. To ensure reproducibility, PyTorch and NumPy were initialized with fixed seeds, and deterministic CUDA settings were enabled during training. A learning rate of $1 \times 10^{-4}$ was used across all seeds and architectures. The final result for each architecture was calculated as the average of the three runs.

AdamW was selected as the optimizer with a weight decay value of $1 \times 10^{-4}$. BCEWithLogitsLoss was employed for model training. The learning rate was adjusted using ReduceLROnPlateau, which dropped it by 50\% when validation loss showed no additional improvement for 2 consecutive epochs. The models were trained for a maximum of 20 epochs. Training was stopped if the validation loss did not improve for 5 epochs. The model with the lowest validation loss was then used for the final test.

\section{Results and Discussion}

All results reported below use a learning rate of $1 \times 10^{-4}$ and are averaged
across three random seeds (9, 99, and 999).

\subsection{Performance with EfficientNet-B0 (Baseline)}

We can see the performance of EfficientNet-B0 in Table~\ref{tab:efficientnet_seeds}.
After 3 seeds, the average accuracy is 84.64\% and ROC-AUC is 92.46\%. The accuracy is
higher than the accuracy of the original dataset paper~\cite{roy2026defactify}, which was
80.14\% using ResNet-50.

\begin{table}[htbp]
\caption{Per-Seed Results — FFT + EfficientNet-B0}
\label{tab:efficientnet_seeds}
\centering
\begin{tabular}{lccccc}
\toprule
\textbf{} & \textbf{Acc} & \textbf{Prec} & \textbf{Rec} & \textbf{F1} & \textbf{AUC} \\
\midrule
Seed 9   & 0.8400 & 0.8242 & 0.8644 & 0.8438 & 0.9210 \\
Seed 99  & 0.8538 & 0.8464 & 0.8644 & 0.8553 & 0.9252 \\
Seed 999 & 0.8453 & 0.8666 & 0.8163 & 0.8407 & 0.9277 \\
\midrule
\textbf{Average} & \textbf{0.8464} & \textbf{0.8457} & \textbf{0.8484} & \textbf{0.8466} & \textbf{0.9246} \\
\bottomrule
\end{tabular}
\end{table}
Our EfficientNet-B0 model improves the accuracy by 4.50
percentage points over the reported accuracy of the dataset paper. At the same time,
EfficientNet-B0 is substantially lighter than ResNet-50, which increases the usability
of the model for lightweight applications.

\subsection{Performance with ECA}

To further investigate the effectiveness of an attention module for detecting AI-generated
images, we used Efficient Channel Attention (ECA). As shown in Table~\ref{tab:eca_seeds},
adding ECA did not improve the model’s performance. It obtained an average accuracy of 84.54\%, slightly below the baseline result of 84.64\%. 
\begin{table}[htbp]
\caption{Per-Seed Results — FFT + EfficientNet-B0 + ECA}
\label{tab:eca_seeds}
\centering
\begin{tabular}{lccccc}
\toprule
\textbf{} & \textbf{Acc} & \textbf{Prec} & \textbf{Rec} & \textbf{F1} & \textbf{AUC} \\
\midrule
Seed 9   & 0.8438 & 0.8485 & 0.8369 & 0.8427 & 0.9230 \\
Seed 99  & 0.8506 & 0.8489 & 0.8531 & 0.8510 & 0.9257 \\
Seed 999 & 0.8419 & 0.8820 & 0.7894 & 0.8331 & 0.9173 \\
\midrule
\textbf{Average} & \textbf{0.8454} & \textbf{0.8598} & \textbf{0.8265} & \textbf{0.8423} & \textbf{0.9220} \\
\bottomrule
\end{tabular}
\end{table}
ECA also lagged behind the baseline in Recall, F1-score, and ROC-AUC. However, Precision
increased from 84.57\% to 85.98\%. Therefore, we can conclude that adding ECA did not
improve the overall performance of the model for AI-generated image detection.

\subsection{Performance with U-Net Decoder}

To evaluate whether a U-Net decoder can improve the frequency features extracted by
EfficientNet-B0, we added a U-Net decoder after the EfficientNet-B0 encoder. As shown
in Table~\ref{tab:unet_seeds}, the U-Net model achieved an average accuracy of 84.85\%,
slightly above the baseline accuracy of 84.64\%. The ROC-AUC also increased
from 92.46\% to 92.77\%. 
\begin{table}[htbp]
\caption{Per-Seed Results — FFT + EfficientNet-B0 + U-Net}
\label{tab:unet_seeds}
\centering
\begin{tabular}{lccccc}
\toprule
\textbf{} & \textbf{Acc} & \textbf{Prec} & \textbf{Rec} & \textbf{F1} & \textbf{AUC} \\
\midrule
Seed 9   & 0.8397 & 0.8852 & 0.7806 & 0.8296 & 0.9201 \\
Seed 99  & 0.8484 & 0.8918 & 0.7931 & 0.8396 & 0.9304 \\
Seed 999 & 0.8575 & 0.8501 & 0.8681 & 0.8590 & 0.9325 \\
\midrule
\textbf{Average} & \textbf{0.8485} & \textbf{0.8757} & \textbf{0.8139} & \textbf{0.8427} & \textbf{0.9277} \\
\bottomrule
\end{tabular}
\end{table}
However, the Recall and F1-score were lower than the baseline.
Precision increased from 84.57\% to 87.57\%. Overall, U-Net provided a small improvement
in Accuracy and ROC-AUC, however the improvement was not seen in every metric.

\subsection{Performance with Attention U-Net Decoder}

Among the tested architectures, Attention U-Net showed the most promising performance.
It achieved an average accuracy of 85.51\%, compared to 84.64\% for the EfficientNet-B0
baseline, which is an improvement of 0.87 percentage points. This is a 5.37 percentage-point improvement over the ResNet-50 baseline reported in the dataset paper. The ROC-AUC also increased
from 92.46\% to 92.99\%. Precision increased from 84.57\% to 87.91\%. The F1-score slightly
improved from 84.66\% to 85.03\%. Table~\ref{tab:attunet_seeds} shows the per-seed
breakdown. Overall, among the tested architectures, Attention U-Net showed a relatively strong overall result. This suggests that adding attention to the U-Net decoder can improve AI-generated image detection in the frequency domain.
\begin{table}[t]
\caption{Per-Seed Results — FFT + EfficientNet-B0 + Attention U-Net}
\label{tab:attunet_seeds}
\centering
\begin{tabular}{lccccc}
\toprule
\textbf{} & \textbf{Acc} & \textbf{Prec} & \textbf{Rec} & \textbf{F1} & \textbf{AUC} \\
\midrule
Seed 9   & 0.8453 & 0.8927 & 0.7850 & 0.8354 & 0.9271 \\
Seed 99  & 0.8588 & 0.8583 & 0.8594 & 0.8588 & 0.9306 \\
Seed 999 & 0.8612 & 0.8864 & 0.8287 & 0.8566 & 0.9320 \\
\midrule
\textbf{Average} & \textbf{0.8551} & \textbf{0.8791} & \textbf{0.8244} & \textbf{0.8503} & \textbf{0.9299} \\
\bottomrule
\end{tabular}
\end{table}

\subsection{Overall Comparison}

The overall performance of the four architectures is compared in
Table~\ref{tab:summary}. EfficientNet-B0 achieved an accuracy of 84.64\% and
ROC-AUC of 92.46\%. ECA did not improve the overall performance, while U-Net provided a
small improvement in accuracy and ROC-AUC. Attention U-Net achieved the highest accuracy
of 85.51\% and the highest ROC-AUC of 92.99\% among the tested architectures. 
It also showed the highest Precision and F1-score. Overall, the results show that adding an attention U-Net decoder provided the best overall performance among the tested approaches.
\begin{table}[h!]
\caption{Overall Comparison of All Architectures (Average over Seeds 9, 99, 999)}
\label{tab:summary}
\centering
\footnotesize
\setlength{\tabcolsep}{4pt}
\begin{tabular}{lccccc}
\toprule
\textbf{Architecture}
  & \textbf{Acc}
  & \textbf{Prec}
  & \textbf{Rec}
  & \textbf{F1}
  & \textbf{AUC} \\
\midrule
EfficientNet-B0         & 0.8464 & 0.8457 & \textbf{0.8484} & 0.8466 & 0.9246 \\
+ ECA                   & 0.8454 & 0.8598 & 0.8265 & 0.8423 & 0.9220 \\
+ U-Net                 & 0.8485 & 0.8757 & 0.8139 & 0.8427 & 0.9277 \\
+ Attention U-Net       & \textbf{0.8551} & \textbf{0.8791} & 0.8244 & \textbf{0.8503} & \textbf{0.9299} \\
\bottomrule
\end{tabular}
\end{table}

\section{Conclusion}

We investigated different approaches to improve AI-generated image detection using frequency-domain representations. EfficientNet-B0 was used as the baseline encoder because of its lightweight architecture, which can provide better deployment capability in resource-limited applications. We further explored whether the extracted frequency features could be improved before classification by using Efficient Channel Attention (ECA). We also tested U-Net and Attention U-Net to study whether decoder-based feature refinement could improve detection performance.

The EfficientNet-B0 baseline achieved an accuracy of 84.64\%, outperforming the ResNet-50 baseline by 4.50 percentage points, which achieved 80.14\% on the MS COCOAI dataset. ECA did not improve the overall performance of the baseline model. U-Net provided a small improvement, increasing the accuracy from 84.64\% to 84.85\% and ROC-AUC from 92.46\% to 92.77\%. Attention U-Net obtained the best overall performance among the tested architectures, with an accuracy of 85.51\%, F1-score of 85.03\%, and ROC-AUC of 92.99\%. Compared with the EfficientNet-B0 baseline, Attention U-Net improved the accuracy by 0.87 percentage points and by 5.37 percentage points compared with the ResNet-50 baseline reported in the dataset paper.

Overall, the results show that decoder-based feature refinement can improve frequency-domain AI-generated image detection. Attention U-Net showed better overall results compared with the other tested models, while adding ECA did not lead to an improvement over the EfficientNet-B0 baseline. These findings suggest that the use of decoder-based feature refinement, particularly Attention U-Net, improves frequency-domain AI-generated image detection. However, the improvement obtained with Attention U-Net over the EfficientNet-B0 baseline was relatively small, indicating that there is still scope for further experimentation. Further work can explore these architectures on different datasets and a broader range of image generation models. Statistical significance testing can also be included to further validate the observed performance differences. The combination of frequency-domain and spatial-domain information can also be investigated to further improve detection performance.

\bibliographystyle{IEEEtran}
\bibliography{Ref}

\end{document}